\documentclass[journal]{IEEEtran}
\ifCLASSINFOpdf
\else
\fi
\usepackage{times}
\usepackage{epsfig}
\usepackage{graphicx}
\usepackage{amsmath}
\usepackage{amssymb}

\usepackage{subfigure}
\usepackage{amsthm}

\usepackage{multirow}
\usepackage{booktabs}
\usepackage{algorithm}
\usepackage{algorithmic}
\usepackage{epstopdf}
\usepackage{color}
\usepackage{hyperref}

\usepackage{makecell}

\begin{document}
%
\title{Simultaneous Region Localization and Hash Coding for Fine-grained Image Retrieval}
%
%
%

\author{Haien Zeng, Hanjiang Lai, Jian Yin
\thanks{Email address: zenghen@mail2.sysu.edu.cn (H. Zeng), (laihanj3, issjyin)@mail.sysu.edu.cn (H. Lai, J. Yin).}
\thanks{H. Zeng, H. Lai, and J. Yin are with School of Data and Computer Science, Sun Yat-Sen University, China. Hanjiang Lai is Corresponding author.}
}

\markboth{}%
{Shell \MakeLowercase{\textit{et al.}}: Bare Demo of IEEEtran.cls for Journals}
%



\maketitle

\begin{abstract}
Fine-grained image hashing is a challenging problem due to the difficulties of discriminative region localization and hash code generation. Most existing deep hashing approaches solve the two tasks independently. While these two tasks are correlated and can reinforce each other. In this paper, we propose a deep fine-grained hashing to simultaneously localize the discriminative regions and generate the efficient binary codes. The proposed approach consists of a region localization module and a hash coding module. The region localization module aims to provide informative regions to the hash coding module. The hash coding module aims to generate effective binary codes and give feedback for learning better localizer. Moreover, to better capture subtle differences, multi-scale regions at different layers are learned without the need of bounding-box/part annotations. Extensive experiments are conducted on two public benchmark fine-grained datasets. The results demonstrate significant improvements in the performance of our method relative to other fine-grained hashing algorithms.
\end{abstract}


%
\IEEEpeerreviewmaketitle

\section{Introduction}
As the explosive increment of images on the Internet, hashing methods~\cite{wang2018survey}, which encode images into binary codes for efficient storage and fast search, have attracted lots of attention from computer vision and multimedia communities. 
Much effort~\cite{lai2018improved,gui2018fast,cao2018deep,lai2019improving} has been devoted to learning similarity-preserving hash functions that map similar/dissimilar samples into nearby/faraway binary codes, e.g., the unsupervised hashing methods~\cite{shen2018unsupervised,liu2017reversed}, the semi-supervised hashing methods~\cite{zhang2017ssdh,SSH} and the supervised hashing methods~\cite{gui2018fast,KSH}.


Most existing hashing models only consider the \textit{coarse-grained} image similarities. That is, when two images belong to a general category, they are considered similar. The coarse-grained image search is not sufficient for real applications. It is also desirable to distinguish categories that are within a general category, e.g, various species of dogs~\cite{KhoslaYaoJayadevaprakashFeiFei_FGVC2011} and birds~\cite{wah2011caltech}. In this paper, we focus on fine-grained hashing.  

Although the existing coarse-grained methods are powerful, there still remains some problems for fine-grained image retrieval. The main difference between the coarse-grained and fine-grained hashing is that the differences of the fine-grained images are more subtle. For example, many previous works, e.g.,~\cite{fu2017look}, indicate that the small local regions play an important role to discriminate the similarities among images, such as the beak or legs for distinguishing the species of birds. Thus it is more challenging to preserve the similarities among fine-grained images compared to the coarse-grained images. And the existing coarse-grained methods always result in suboptimal solutions for fine-grained databases because they are not designed to capture the subtle differences. 

Limited attention has been paid for fine-grained hashing. Deep saliency hashing~\cite{jin2018deep} uses the attention mechanism to learn the fine-grained hashing codes, in which the attention map is firstly learned and then the binary codes are generated lately. Feature pyramid hashing~\cite{FPH} is a two-pyramid hashing network for fine-grained retrieval, in which the vertical pyramid is to capture the semantic differences and the horizontal pyramid is proposed for subtle differences. However, these methods perform the region localization and hash coding independently, leading to the suboptimal localizer and hash functions.


Specifically, the discriminative parts of the fine-grained objects are always very small, thus it is hard to localize these regions. Due to the inaccuracy regions, it is also difficult to generate effective binary codes. Hence, the following two problems of the fine-grained hashing should be simultaneously addressed: 1) discriminative region localization and 2) hash code generation.

\begin{figure}[t]
  \centering
    \includegraphics[width=0.88\hsize \hspace{0.01\hsize}]{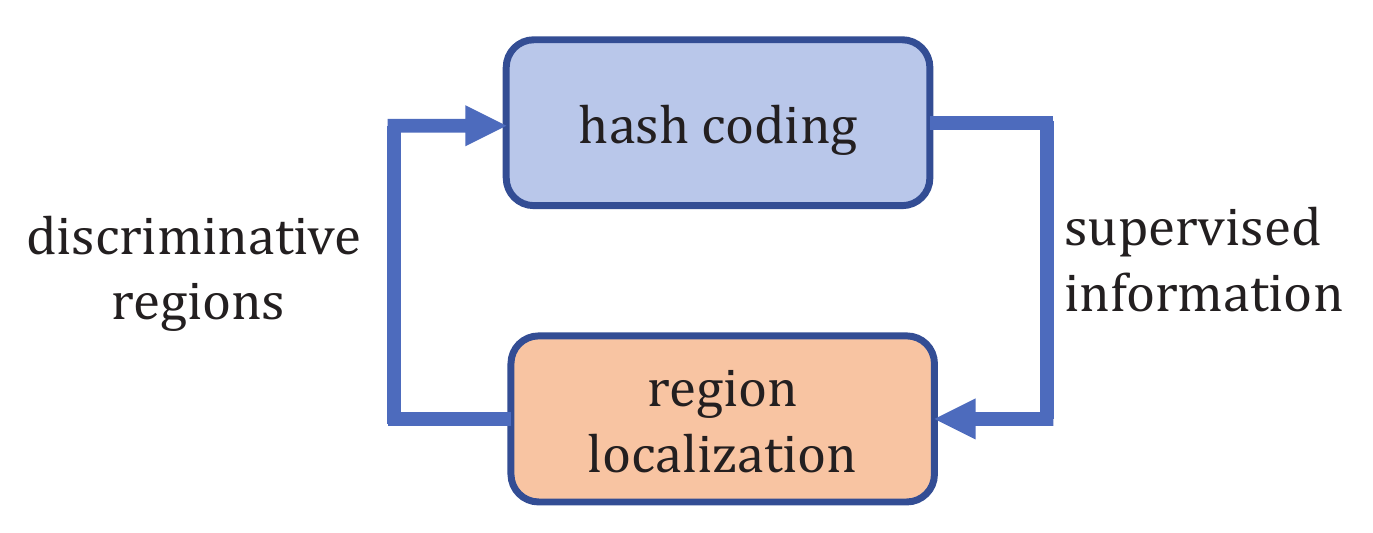}
  \caption{Illustration of our method. It consists of two blocks: region localization and hash coding modules. The hash coding module encodes the whole image and multiple informative regions into the efficient binary codes, and also provides the supervised information to the region localization module. The region localization module outputs discriminative regions to the hash coding module.  The hash coding and region localization are trained simultaneously. }
  \label{motivation}  
\end{figure}

It is challenging to find discriminative regions that are favorable for retrieval. One solution is to localize informative regions by leveraging the supervised bounding-box/part annotations~\cite{huang2016part}. While the supervised part detectors require the great cost on human annotation. It makes these approaches not practical for large-scale fine-grained image retrieval problem. Hence, some~\cite{dubey2018pairwise} try to find informative regions without the need of the ground-truth bounding-box/part annotations. However, these unsupervised methods do not consider to discriminate the subtle similarities among fine-grained images, which is very important for similarity-preserving hashing.  

In this paper, we propose a fine-grained hashing to solve the two tasks without the need of bounding-box/part annotations. The main assumption behind our method is that the region localization and hash coding are mutually correlated and can be learned in a reinforced way. For region localization, the main problem is that we do not have the ground-truth annotations. To solve this, we use the information from the hash coding module to guide the training of the localizer, that is \textit{the regions that achieve better hashing performance are the more discriminative regions}. For hash coding, the main problem is that we do not have the discriminative regions. It can be easily provided by the region localization module. Observed by that, our method consists of two building blocks: discriminative region localization and hash coding. As shown in Figure~\ref{motivation}, the two building blocks are trained simultaneously. First, the region localization module finds informative regions with various scales for the hash coding module. Given the informative regions from the region localization module, the hash coding module then is learned to generate better binary codes, which also provides the supervised information, e.g., which regions achieve better performances, to the region localization module. Knowing the performances of all regions, we can retrain the region localization module to obtain more discriminative regions. 

More specifically, the proposed method is illustrated in Figure~\ref{collaborative_learning}. The hash coding module first takes several informative regions as input. And then a ranker network and a comparer network, which use multiple intermediate features to learn better hash codes, are proposed. Finally, the hash codes are learned by minimizing the proposed objective function defined over classification error and ranking error, in which the characteristics of classification and retrieval are both obtained in our model. On the region localization module, we firstly use the comparer network to find the most discriminative regions. Multiple attention regions at different layers are found without the need of bounding-box/part annotations. We use such supervised information to retrain the region localization module. 
The collaborative learning mechanism is used to learn both the region localization and hash coding. Two modules can complement and reinforce each other.


The main contributions of this paper are summarized as follows.
\begin{itemize}
\item A novel hashing method is designed for fine-grained image retrieval, which simultaneously localizes discriminative regions and generates effective binary codes in a mutually reinforced way.


\item A collaborative learning mechanism is used to optimize the region localization module and the hash code generation module. Such a design enables our model to localize the fine-grained regions without the need of bounding box/part annotations.

\item The comprehensive experiments are conducted on two benchmark datasets (Stanford Dogs and CUB Birds). The results show that the proposed method achieves superior performance over the state-of-the-art baselines.

\end{itemize}

\section{Related Work}

\subsection{Hashing}
Hashing is a popular method for nearest neighbor search. Learning-based hashing methods~\cite{norouzi2011minimal,liu2012supervised} with a shallow architecture are firstly proposed in the literature. For instance, locality sensitive hashing (LSH)~\cite{gionis1999similarity} uses a random hyperplane to encode the data into binary codes. It is the most popular unsupervised method. Iterative quantization (ITQ)~\cite{gong2013iterative} finds a rotation of zero-centered data to minimize the quantization error. Fast supervised discrete hashing (FSDH)~\cite{gui2018fast} uses the class labels to learn the corresponding hash codes, which is scalable to the large-scale database. 

Furthermore, the deep-network-based hashing methods~\cite{lai2015simultaneous,cao2017hashnet,liu2017deep,qiu2017deep,gui2018fast,wang2018survey,cao2018deep} have been shown much better performance for image retrieval since the powerful image representations and hash-code learning can be simultaneously learned in the deep networks. Deep hashing methods~\cite{wang2018survey} learn similarity-preserving hash functions based on deep neural networks, in which convolutional neural network (CNN) is the most widely used neural architecture for image retrieval. Xia et al.~\cite{xia2014supervised} proposed a deep supervised hashing method, in which the image representations as well as a set of hash functions are automatically learned. Following this work, many deep CNN-based hashing methods are proposed. For example, deep pairwise-supervised hashing (DPSH)~\cite{Li2016Feature} and deep supervised hashing (DSH)~\cite{liu2016deep} are two representative pairwise methods to learn the binary codes. Deep cauchy hashing~\cite{cao2018deep} is proposed to enable efficient and effective Hamming space retrieval, where a pairwise cross-entropy loss based on Cauchy distribution is proposed. Some deep learning methods~\cite{lai2015simultaneous,lai2016instance} preserve relative similarity relations via triplet labels.  Deep triplet quantization~\cite{liu2019deep} is a novel approach to learn deep quantization model from the similarity triplets. HashNet~\cite{cao2017hashnet} is proposed to directly optimize deep networks with non-smooth binary activation.  

Subsequently, inspired by the success of the generative neural network (GAN)~\cite{goodfellow2014generative}, some recent research has employed GANs to image retrieval. For example, DSH-GAN~\cite{qiu2017deep} is proposed for semi-supervised hashing. The synthesized images generated by GAN are utilized to produce better binary codes. HashGAN~\cite{cao2018hashgan} learns compact binary hash codes from both real image and synthesized images. Zhang et al.~\cite{zhang2018attention} proposed an attention-aware hashing method for cross-modal hashing. Wang et al.~\cite{wang2017adversarial} present an adversarial cross-modal retrieval method that seeks a common subspace based on adversarial learning. Very recently, some effort has been made in hashing methods with reinforcement learning, e.g.,~\cite{zhang2018deep,lai2019}.

Limited attention has been paid for fine-grained hashing. Jin et al.~\cite{jin2018deep} proposed deep saliency hashing for fine-grained hashing, which finds salient regions to learn binary codes. Yang et al.~\cite{FPH} used the low-level feature activations to capture the subtle differences. Existing approaches mainly solve the discriminative region localization and hash code generation independently, while neglecting the discriminative region localization and hash code generation are dependent and should be learned simultaneously.

\begin{figure*}[t]
  \centering
    \includegraphics[width=1\hsize \hspace{0.01\hsize}]{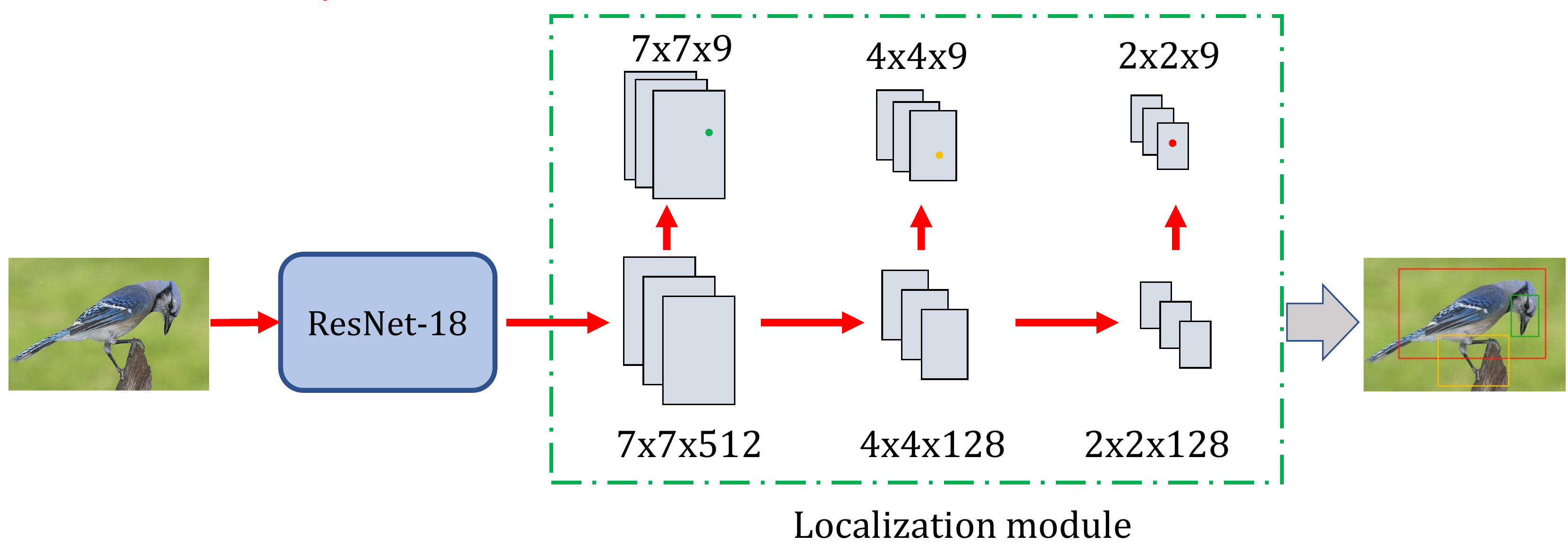}
  \caption{Overview of the region localization module. It consists of two sequential parts:  feature extractor (ResNet-18) and localization. In this paper, we use multi-scale feature maps of sizes $\{7 \times 7, 4 \times 4, 2 \times 2\}$ from  three different layers, in which the feature map pixels in the earlier layer correspond to smaller regions. Since we use three scales and three ratios, it yields  $9 = 3 \times 3$ proposals at each feature map pixel. The main problem is that we do not have the annotations.
  }
  \label{localization}
\end{figure*}

\subsection{Discriminative Region localization}
There are two main categories to localize regions for fine-grained recognition: 1) supervised approaches~\cite{huang2016part,lin2015deep} that leverage the bounding-box/part annotations to find the discriminative region and 2) unsupervised approaches~\cite{fu2017look,dubey2018pairwise} that learn visual attention maps without the annotations. For supervised approaches, many methods have been proposed. For example, Huang et al.~\cite{huang2016part} proposed a part-stacked CNN architecture to model subtle differences using manually-labeled annotations. Semantic part detection and abstraction(SPDA-CNN)~\cite{zhang2016spda} has two sub-networks, i.e., detection and recognition, for fine-grained classification. Since it is time-consuming to collect enough manual annotations, the supervised approaches may be not practical for large-scale databases. Thus attention-based approaches have been proposed to find regions without using any annotations. Fu et al.~\cite{fu2017look} proposed a recurrent attention convolutional neural network (RA-CNN) to jointly learn discriminative region and feature representation at multiple scales. A self-supervision mechanism~\cite{yang2018learning} is proposed to localize informative regions without bounding-box/part annotations. Sun et al.~\cite{sun2018multi} applied the multi-attention multi-class constraint to learning the correlations among object parts.

These methods can find the discriminative region localization, but they do not consider how to encode these subtle differences into binary codes. In this work, we propose a fine-grained hashing to address the two challenges in a mutually reinforced way.



\section{Our Method}
Suppose that we have a set of $n$ training images $S = \{I_i, y_i\}_{i=1}^n$, where $I_i$ is the $i$-th image, $y_i$ is one of $C$ fine-grained labels, $y_i = 1, \cdots, C$. The goal of fine-grained hashing is to learn similarity-preserving hash functions that map the fine-grained images into binary codes. And the similarities are also preserved in the Hamming space. Let $H_i \in \{-1,1\}^{b}$ is the $b$-dimensional binary code associated with the $I_i$ in Hamming space. 

In this section, we present the fine-grained hashing without the bounding-box/part annotations, which consists of two building blocks: one is region localization and another is hash code generation. We firstly  introduce the two modules separately. The problems of the two modules are also shown. Finally, we propose a collaborative learning mechanism to simultaneously optimize the two modules.



\subsection{Discriminative Region Localization} \label{region_localization}

 
For a deep convolutional neural network, there are many layers. Each layer has the corresponding feature maps. A feature map pixel has a receptive field on the image, e.g., the SPP-net~\cite{lenc2015r} maps a feature map pixel to the center of the receptive field on the image.  The feature map size always decreases with the increase in the layers. Thus, for the lower layer, a feature map pixel has a smaller receptive field. The higher layer has a larger receptive field. Similar to region proposal network (RPN)~\cite{ren2015faster}, we use three different layers with feature map sizes $\{7 \times 7, 4 \times 4, 2 \times 2 \}$ to capture different scales of the objects. We also use three sizes $\{32, 48, 96\}$ and three ratios $\{1:1, 2:3, 3:2\}$ to generate region proposals. 

\begin{figure*}[t]
  \centering
    \includegraphics[width=1\hsize \hspace{0.01\hsize}]{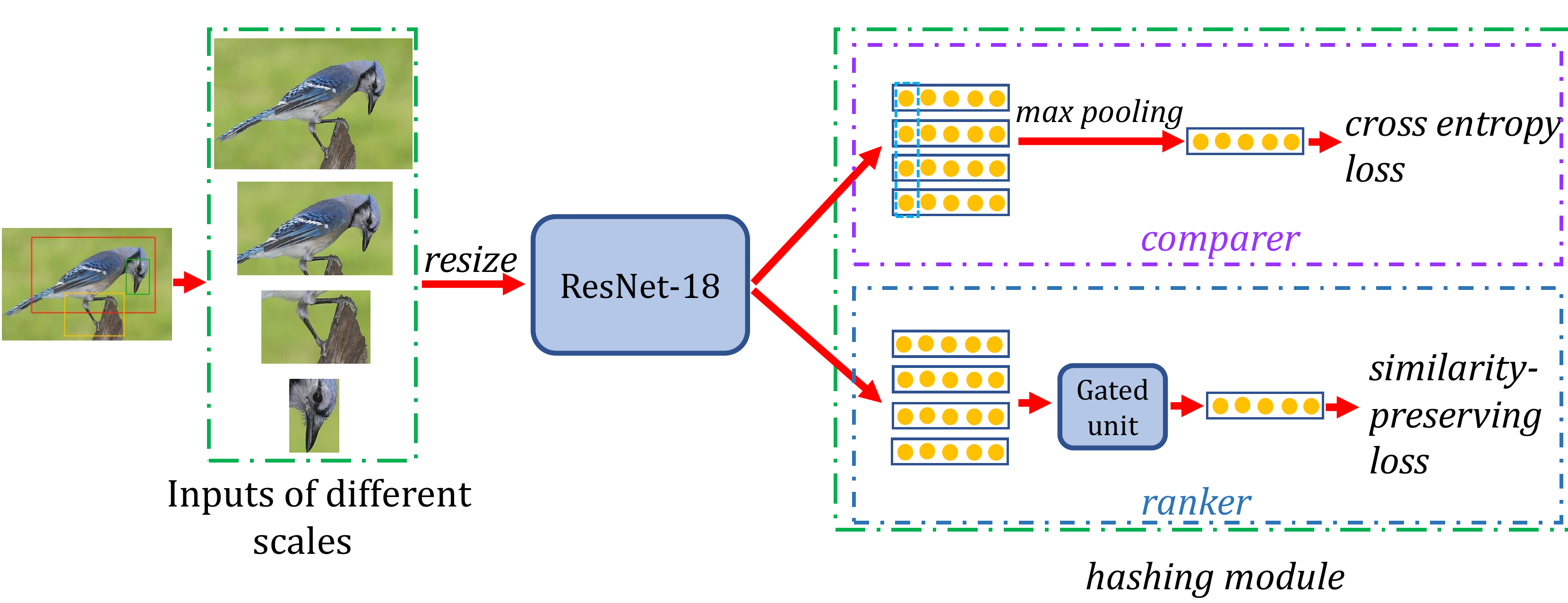}
  \caption{Overview of the hash code generation module. It consists of four parts: input layer, feature extractor (ResNet-18), comparer and ranker. The comparer aims to find the most discriminative proposal. The ranker combines all features to generate better binary codes. The main problem is that we do not have the informative regions as inputs.
  }
  \label{hash_coding}
\end{figure*}

Figure~\ref{localization} gives the specific structure of the region localization network. Given an input image of size $224 \times 224$, it goes through the ResNet-18~\cite{he2016deep} network. The feature maps of sizes $\{7 \times 7, 4 \times 4, 2 \times 2 \}$ are constructed. Hence, three different layers with feature map sizes $\{7 \times 7 \times 512, 4 \times 4 \times 128, 2 \times 2 \times 128 \}$ are added on the top of ResNet-18. As suggested in RPN~\cite{ren2015faster}, we also use three scales and three ratios to capture different sizes of the regions. That is, it yields $R = 9$ proposals at each feature map pixel. To assign scores to these 9 proposals, a convolutional layer with kernel size $1 \times 1$ is added on the top of each layer, respectively. And the outputs are $\{7 \times 7 \times 9, 4 \times 4 \times 9, 2 \times 2 \times 9 \}$. In summary, we generate $7 \times 7 \times 9 + 4 \times 4 \times 9 + 2 \times 2 \times 9 = 621$ region proposals in total.  We make the feature maps as the prediction scores for all region proposals. The three outputs in Figure~\ref{localization} are the scores for these region proposals, in which the larger value, the more informative proposal. In this work, we only select one region proposal that has the largest score in each layer, that is the largest value in each output will be chosen. Since we use multiple scales and ratios, the three most informative regions with different sizes will be selected.



Now, the problem becomes how to learn the region localization module. However, we do not have the annotations of the bounding box. Without supervised information, the region localization module cannot be directly learned. Thus, the first problem arises:

 \textit{How to optimize the region localization module without the bounding-box/parts annotations?}

\subsection{Hash Code Generation}
\label{hash_coding_section}
Most existing deep hashing approaches only take the whole image as input. However, the discriminative regions of the fine-grained objects are always very small. It is difficult for the existing coarse-fined hashing methods to capture them. Thus it is also hard to encode the subtle differences into binary codes. One possible solution is to take the informative regions as inputs to the hashing network. Thus the second problem arises: 

\textit{How to localize the informative regions}? 

Suppose that we can localize the informative regions, e.g., the outputs of the region localization network are the three informative regions with different scales. Please note that how to obtain these informative regions will be described in the next subsection. An illustration of the deep network for hash code generation is shown in Figure~\ref{hash_coding}. The network is divided into four  parts: 1) the input layer which has several informative regions with different scales and one whole image; 2) the feature extractor that maps all input images into efficient feature representations; 3) the comparer which compares the regions and tells which region is most informative and 4) the ranker which includes a gated network and a similarity-preserving loss to learn the binary codes. The gated network fuses all features into a joint representation and the similarity-preserving loss aims to learn the good hash functions.

\textbf{Input layer} Since the discriminative regions are always very small, we zoom in the selected regions according to~\cite{fu2017look}, which shows that the subtle difference can be more significant when the attended regions are zoomed at a finer scale. Specifically, we crop the regions from the full image $I_0$, and resize them to the size of $224 \times 224$ (the input size of the feature extractor) with higher resolution. Let $I_1, I_2$ and $I_3$ denote the cropped images for the three regions. 
The network takes four images as inputs: $I_0$, $I_1, I_2$ and $I_3$. Please note that how to obtain the discriminative regions is described in Subsection~\ref{training_hash_coding}.

\textbf{Feature Extractor} We use ResNet-18 as the basic architecture to learn efficient image features. ResNet-18 is a residual learning framework and has shown its success in many computer vision tasks. It can generate powerful image representations for hashing. We make the following structural modification. We use all layers until the Conv5\_2 layer in ResNet-18, and the last 1000-way fully connected layer is removed. The feature maps in Conv5\_2 are used as the image intermediated features. We denote $F_0, F_1, F_2$ and $F_3$ as the intermediated feature maps for the four inputs, respectively. 

Since $F_i$ is feature maps, the global average pooling (GAP) is used to reduce the feature maps to feature vector by taking the average of each feature channel. Let $f_0, f_1, f_2$ and $f_3$ denote the output vectors of the GAP layer associated with $F_0, F_1, F_2$ and $F_3$, respectively.

\textbf{Comparer} The comparer aims to tell which region is most informative. Suppose that there are $C$ labels in total, all $N=4$ feature vectors, i.e., $f_0, f_1, f_2$ and $f_3$, go through a single-layer neural network with a fully-connected layer that maps the feature vectors to the $C$-dimensional vectors. After that, we get a matrix $P$. The $P \in \mathbb{R}^{N \times C}$ represents the label probability matrix, where $P(i,:)$ is the $C$-dimensional vector for the $i$-th proposal.

While, we have the ground truth labels for the whole image $I_0$ but do not have the class labels for each proposal. How to use the three region proposals? In this paper, we propose a comparative procedure. For example, when the $i$-th region proposal is more discriminative than the $j$-th proposal for the fine-grained label $c$, then the value of $P(i,c)$ should be larger than that of $P(j,c)$. Hence, when $P(i,c)$ is the largest value in the $c$-th row, then the $i$-th proposal is the most discriminative for the fine-grained label $c$. Inspired by that, we use a max-pooling layer to find most discriminative proposal. We fuse $P(1,:), P(2,:), \cdots, P(N,:)$ into one $C$-dimensional vector via a simple max pooling operator. For the $c$-th column, we find the maximum in the matrix $P$ as
\begin{equation}
\begin{aligned}
p(c) = & \max\{P(1,c), P(2,c), \cdots, P(N,c)\},& \\ &\forall c = 1, \cdots, C,&
\end{aligned}
\end{equation}
where $p \in \mathbb{R}^C$ is a $C$-dimensional vector, and $p(c)$ is the maximum value in the $c$-th column of the matrix $P$.  The $p$ is forwarded to a softmax layer to obtain a probability distribution $m \in \mathbb{R}^C$ by
\begin{equation}
m(c) = \frac{\exp(p(c))}{\sum_{i=1}^C \exp(p(i))},
\end{equation}
where $m(c)$ can be regarded as the probability score that the all proposals contains a fine-grained object in the $c$-th category. When the $j$-th proposal contains the $c$-th category, then the probability of $m(c)$ should be a large value and $P(j,c)$ will have a high response. Hence, it can guide the learning of comparer that is able to find the most discriminative region proposal. 

Given the ground-truth label $y$, it becomes a traditional image classification problem. A simple softmax loss function~\cite{krizhevsky2012imagenet} can be used to train the network, which is formulated as
\begin{equation}
\min - \log(\frac{m(y)}{\sum_{c=1}^C m(c)}).
\end{equation}

Please note that it is widely used loss function for classification. Recently, some hashing methods have been proposed to optimize the objective function that defined over both the classification error and the ranking error~\cite{yang2018supervised}, and they showed that these approaches can enhance the deep architecture for hashing. In this work, we add such classifier to learn better deep network. Moreover, this loss function also can provide the information to the region localization module, e.g., which is the most discriminative proposal?

After training the network, the region proposals that can capture the subtle differences will have higher responses. In the next subsection~\ref{training_localization}, we will show how to use the comparer network to select the most informative region proposal.

\textbf{Ranker} In the other hand, we have three features that extracted from three regions ($f_1,f_2,f_3$) and one feature from the whole image ($f_0$). Now, we aim to combine these features to a more powerful joint feature and more efficient binary codes. 

Recently, several algorithms have been proposed based on gated neural network for the combination of data from multiple inputs, e.g., the gated multimodal unit (GMU)~\cite{arevalo2017gated}. The gated network likes LSTM or GRU, in which a gate neuron (e.g., a sigmoid activation function) controls the contribution of the features and is to decide whether the features may contribute or not. To generate a better joint representation, in this work, we use the gated neural network to automatically decide which inputs are more likely to contribute to correctly generate binary codes.

Concretely, the four features ($f_0, f_1, f_2, f_3$) further go through a fully-connected layer to produce four intermediate image representations: $\hat{f}_0, \hat{f}_1, \hat{f}_2$ and $\hat{f}_3$, respectively. Then these four features are fed to the gated network to selectively find the informative information for encoding the binary codes. The gated network takes $\hat{f}_0, \hat{f}_1, \hat{f}_2$ and $\hat{f}_3$ as inputs, and $h$ as output:


\begin{equation}
\begin{aligned}
 &h_0 = \tanh(W_0 \hat{f}_0 + b_0)& \\
 &h_1 = \tanh(W_1 \hat{f}_1 + b_1)& \\
 &h_2 = \tanh(W_2 \hat{f}_2 + b_2)& \\
 &h_3 = \tanh(W_3 \hat{f}_3 + b_3)& \\
 &C = [\hat{f}_0;\hat{f}_1;\hat{f}_2;\hat{f}_3]& \\
 &z_0 =\sigma(W_{z_0}C + b_{z_0})&\\
 &z_1 =\sigma(W_{z_1}C + b_{z_1})&\\
 &z_2 =\sigma(W_{z_2}C + b_{z_2})&\\
 &z_3 =\sigma(W_{z_3}C + b_{z_3})&\\
 &h = h_0*z_0 + h_1*z_1 + h_2*z_2 + h_3*z_3,&
\end{aligned}
\end{equation}
where $W_0, W_1, W_2$ and $W_3$ are transformation matrices, and $b_0, b_1, b_2$ and $b_3$ are model biases. The tanh represents a tanh activation function. $C$ is the vector that concatenates all the four features. The $W_{z_0}, W_{z_1}, W_{z_2}$, $W_{z_3}$ are transformation matrices, and $b_{z_0}, b_{z_1}, b_{z_2}$, $b_{z_3}$ are biases.  And $\sigma$ denotes a sigmoid activation function. All transformation matrices and biases are needed to be learned.

After the four features are forwarded to the gated network, we can obtain a more powerful and joint representation $h$ that combines all useful information contained in the four features. With the joint representation $h$, it is forward to a $b$-way fully connected layer to generate $b$-bit binary code $H$. Finally, a similarity-preserving loss function is defined by
\begin{equation}
\begin{aligned}
& \sum_{\langle i, j, k \rangle} \max\{0, \varepsilon + ||H_i - H_j|| - ||H_i - H_k||\}, \\
\end{aligned}
\end{equation}
where $\langle i, j, k \rangle$ is the triplet form for three images $I_i, I_j$ and $I_k$, in which the image $I_i$ is more similar to $I_j$ than to $I_k$. And $H_i, H_j$ and $H_k$ are the outputs of the hashing network associated with the $I_i, I_j$ and $I_k$, respectively.

\subsection{Collaborative Learning}
The region localization and the hash code generation cannot be learned separately because of the above two problems, i.e., 1) there are no annotations for learning region localization and 2) the informative regions are unknown for learning the hash codes. Fortunately, the hash coding module can provide annotations to the region localization module and the region localization module can give the informative regions to the hash coding module. Thus, we propose a collaborative learning mechanism to simultaneously optimize these two modules.

\begin{figure}[t]
  \centering
    \includegraphics[width=1\hsize \hspace{0.01\hsize}]{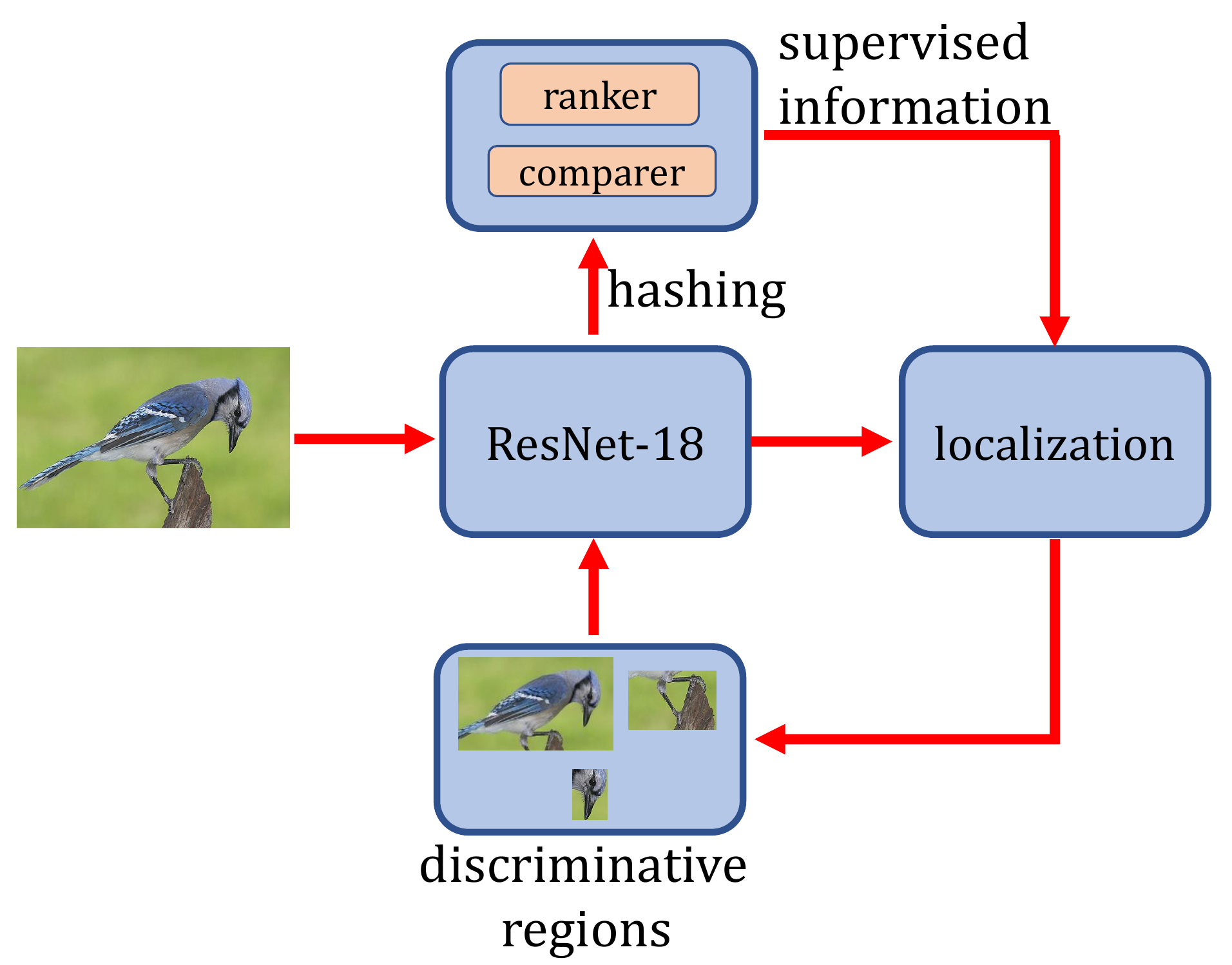}
  \caption{Collaborative learning of two modules. The hash coding and localization are trained in a unified network and in a mutually reinforced way.
  }
  \label{collaborative_learning}
\end{figure}

Figure~\ref{collaborative_learning} shows the proposed deep architecture. First, since learning powerful image representations is the critical component for both the two tasks, we propose a unified feature extractor, i.e., ResNet-18, for the two modules. Collaborative learning is described below.

\subsubsection{Training Region Localization Module with the Help of Hash Coding Module} \label{training_localization}

As shown in Subsection~\ref{region_localization}, an input image goes through the region localization, and the outputs are $\{7 \times 7 \times 9, 4 \times 4 \times 9, 2 \times 2 \times 9 \}$. We aim to select one region proposal for each layer. That is, we only choose 3 most discriminative proposals. In such setting, we can use the multi-scale regions to capture the subtle differences in different sizes. 

Formally, we make each feature maps to predict one region proposal as follow: the most discriminative proposal has the largest value in the feature maps. Taken $7 \times 7$ as an example, let the feature map size to be $H \times W$ (i.e., $H = W =7$) and $R=9$ that is the number of proposals for each feature map pixel, we denote the  feature maps as $A \in \mathbb{R}^{H \times W \times R}$ and the $A(i,j,r)$ is the score for the corresponding region proposal in the $r$-th feature map. And $G$ denotes as the generated region proposals, where $G_{i \times j \times r}$ is the $i \times j \times r$-th region proposal associated with $A(i,j,r)$. The values in feature maps are used as the prediction scores for all region proposals. The larger values in feature maps, the more discriminative region proposals. For example, if $A(i,j,r) < A(h,w,r)$, then the region proposal $G_{h \times w \times r}$ is more discriminative than the $G_{i \times j \times r}$. Suppose that the $h \times w \times r$-th region proposal is the most discriminative proposal, the problem can be defined as: we need to maximize the value of $A(h,w,r)$, that is $A(h,w,r)$ is larger than others. And the objective can be formulated as
\begin{equation}
\min \sum_{\langle i, j, l \rangle \& \langle i, j, l \rangle \neq \langle h, w, r \rangle} \max(0, \epsilon + A(i,j,l) - A(h,w,r)),
\label{objective_localization}
\end{equation}
where $\epsilon$ is the margin, $\langle i, j, l \rangle$ is the indexed of the tensor $A$ where $i=1,\cdots,H, j=1,\cdots,H,$ and $l=1,\cdots,R$. The above loss function requires the value of $A(h,w,r)$ is larger than other scores. With this, we can choose the maximum value as the most discriminative region.

However, we do not have the bounding-box/part annotations, thus the $\langle h, w, r \rangle$ is unknown. In the hash coding module, we have a  comparer, in which we use a comparative procedure to compare the region proposals. The informative region proposals are likely to obtain higher confidence scores on the ground-truth class label. That is, all region proposals go through the comparer, the most informative regions will have the largest score.

\begin{table*}[htbp]
\caption{MAP of Hamming ranking w.r.t different number of bits on two fine-grained datasets.}
\label{MAP}
\centering
\begin{tabular}{|c|
p{1.2cm}<{\centering}p{1.2cm}<{\centering}p{1.2cm}<{\centering}p{1.2cm}<{\centering}|
p{1.2cm}<{\centering}p{1.2cm}<{\centering}p{1.2cm}<{\centering}p{1.2cm}<{\centering}|
}
\hline
\multirow{2}{*}{{Methods}}& \multicolumn{4}
{c|}{CUB-200-2011}&\multicolumn{4}
{c|}{Stanford Dogs}\\
& 16bits & 32bits & 48bits & 64bits
& 16bits & 32bits & 48bits & 64bits
\\
\hline
{Ours}&  {\bf 0.6908}&  {\bf 0.6922}&  {\bf 0.7087}& {\bf 0.7010}& {\bf
0.6668}&  {\bf 0.7402}&  {\bf 0.7516}& {\bf 0.7538}\\ \hline

{FPH}&  {0.5169}&  {0.5832}&  {0.6124}& {0.6233}& {0.6340}&  {0.6909}&  {0.7060}& {0.7130}\\ \hline

DTH &  0.4641&  0.5454&  0.5771&  0.5881&    0.5435&  0.6258&  0.6362&  0.6573\\  \hline

DSH &  0.3156&  0.4930&  0.5408&  0.5967&    0.4728&  0.5587&  0.6128&  0.6319\\  \hline
HashNet &  0.3791&  0.4628&  0.4853&  0.5123&    0.4745&  0.5521&  0.5575&  0.5934\\  \hline

DPSH &  0.3497&  0.4301&  0.4908&  0.5225&    0.4270&  0.5528&  0.6080&  0.6231\\  \hline \hline

CCA-ITQ&      0.1142&  0.1580&  0.1813&  0.1986&    0.2632&  0.3681&  0.4175&  0.4402\\  \hline

MLH&      0.0915&  0.1289&  0.1281&  0.1983&    0.2735&  0.3531&  0.3831&  0.4084\\  \hline

ITQ&      0.0637&  0.0907&  0.1048&  0.1129&    0.2023&  0.2838&  0.3123&  0.3248\\  \hline
SH&      0.0453&  0.0595&  0.0643&  0.0686&    0.1362&  0.1628&  0.1859&  0.1832\\  \hline
LSH&      0.0162&  0.0234&  0.0302&  0.0340&    0.0297&  0.0517&  0.0640&  0.0850\\  \hline

\end{tabular}
\end{table*}

Specially, given a feature map of size $H \times W$, we can generate $N = H \times W \times R$ region proposals. Since $N$ is always large, we adopt non-maximum suppression (NMS) on the regions based on their confidence scores. Then top-$\hat{N}$ region proposals go through the comparer network, and we get the confidences $P$. After that, we find the largest value in the $y$-th column of $P$ as
\begin{equation}
c = argmax \{P(1,y), P(2,y), \cdots, P(\hat{N},y) \},
\end{equation}
where $y$ is the fine-grained class label. Then the $c$-th proposal is the most informative proposal. Knowing $c$, we can easily find the $\langle h, w, r \rangle$ via $(r-1) \times (H \times W) + (w-1) \times H + h = c$. 

With this supervised information from the hash coding module, we can retrain the region localization module using E.q.~\ref{objective_localization}.


\subsubsection{Training Hash Coding Module with the Help of Region Localization Module}
\label{training_hash_coding}

To train the hash coding module, we use the region localization module to find the informative regions. The fine-grained image goes through the localization module, we get three feature maps $\{7 \times 7 \times 9, 4 \times 4 \times 9, 2 \times 2 \times 9 \}$ for different scales. As mentioned above, we only select one proposal for each layer. For each feature maps, the largest confidence score is chosen, and then the associated region proposal is selected as the informative proposal. 

Formally, let $A_1 \in \mathbb{R}^{7 \times 7 \times 9}, A_2 \in \mathbb{R}^{4 \times 4 \times 9}, A_3 \in \mathbb{R}^{2 \times 2 \times 9}$ represent the three feature maps, respectively. Then the largest values in the three tensors can be found as
\begin{equation}
\begin{aligned}
&\langle h_1, w_1, r_1 \rangle = argmax \{ A_1 \},& \\
&\langle h_2, w_2, r_2 \rangle = argmax \{ A_2 \},& \\
&\langle h_3, w_3, r_3 \rangle = argmax \{ A_3 \}.& \\
\end{aligned}
\end{equation}

Then the three most informative region proposals with different scales can be obtained: $G^1_{ h_1 \times w_1 \times r_1}$, $G^2_{ h_2 \times w_2 \times r_2}$ and $G^3_{h_3 \times w_3 \times r_3}$, where $G^1, G^2$ and $G^3$ are the region proposals associated with $A_1, A_2$ and $A_3$. We can crop the regions from the image and get $I_1, I_2$ and $I_3$, respectively. Please note that $I_1, I_2$ and $I_3$ are the inputs to the hash coding module. After that, we can train the hash coding module, which is shown in Subsection~\ref{hash_coding_section}.

\textbf{Discussion} Please note that the comparer can also find the discriminative regions. But the comparer is not utilized to find the informative regions when querying because it is very time-consuming, in which all region proposals are needed to be fed into the comparer network. It is not practical since the number is always a large value, e.g., $7 \times 7 \times 9 + 4 \times 4 \times 9 + 2 \times 2 \times 9 = 621$.

\section{Experiments}
In this section, we extensively evaluate and compare the performance of the proposed method with several state-of-the-art baselines.

\subsection{Databases}
Two benchmark fine-grained datasets are used to evaluate the performance: CUB-200-2011~\cite{wah2011caltech} and Standford Dogs~\cite{KhoslaYaoJayadevaprakashFeiFei_FGVC2011}. These two databases are challenging datasets for image retrieval due to the subtle inter-class difference and large intra-class variation.

\begin{itemize}
\item \textbf{CUB-200-2011}~\footnote{http://www.vision.caltech.edu/visipedia/CUB-200-2011.html}: This dataset is an extended version of CUB-200, which includes 200 bird species. The total number of images is 11,788. For a fair comparison, the official train/test partitions are utilized to build the query set and the retrieval database. Especially, the 5,749 test images are used as the query set. The 5,994 training images are used as the retrieval set. We also utilize the 5,994 training images as the training set.

\item \textbf{Standford Dogs}~\footnote{http://vision.stanford.edu/aditya86/ImageNetDogs/}: This dataset consists of 20,580 images of 120 breeds of dogs from around the world. The images were downloaded from ImageNet. We also utilize the official split to construct the query and retrieval sets. The 8,580 test images and the 12,000 training images are used as the query set and training set, respectively. The 12,000 training images are also used as the retrieval database.
\end{itemize}

For a fair comparison, all of the methods use identical retrieval, training, and test sets.

\subsection{Evaluation Measures}
Four evaluation metrics are used to measure the performance: mean average precision (MAP), precision-recall curves, precision curve within Hamming radius $3$ and precision curve w.r.t. different numbers of top returned instances.

Precision and recall are easily calculated, in which the precision is the fraction of retrieved samples that are relevant and recall is the fraction of relevant samples that are retrieved. MAP is the most widely used evaluation measure for hashing, which is defined as
\begin{equation}
MAP =  \frac{1}{n_q} \sum_{i=1}^{n_q}  \frac{1}{n^i_+} \sum_{k=1}^{n} P_k \times pos_k,
\end{equation}
where $n_q$ is the number of queries. For each query, we compute an average precision score. The average of all queries' average precision scores is the value of MAP. To compute the average precision score, $n^i_+$ denotes the number of relevant samples in the $i$-th ranking list for the $i$-th query, and $P_k$ is the precision score at top-$k$ returns. The $pos_k$ is an indicator, in which $pos_k = 1$ if the $k$-th returned image is relevant to the query, otherwise $pos_k = 0$.


\subsection{Experimental Settings}
For a fair comparison, ResNet-18~\cite{he2016deep} is adapted as the feature extractor for all the deep-network-based methods. Note that the last global average pooling layer and the 1000-way fully connected layer are removed. We use the pre-trained ResNet-18 model that learns from the ImageNet dataset to initialize the feature extractor. The images are resized into $224 \times 224$. In all experiments, we train the networks by the stochastic gradient solver, i.e., ADAM. The batch size is 50, and the base learning rate is 0.0001, which is changed to one-tenth of the current value after every 100 epochs. The weight decay parameter is 1e-5. And $\hat{N}=6$ in our method.

\begin{figure*}[ht!]
\centering
\subfigure[Precision within Hamming radius 3]
{
\label{topp:subfig:a}
\includegraphics[width=2.15in]{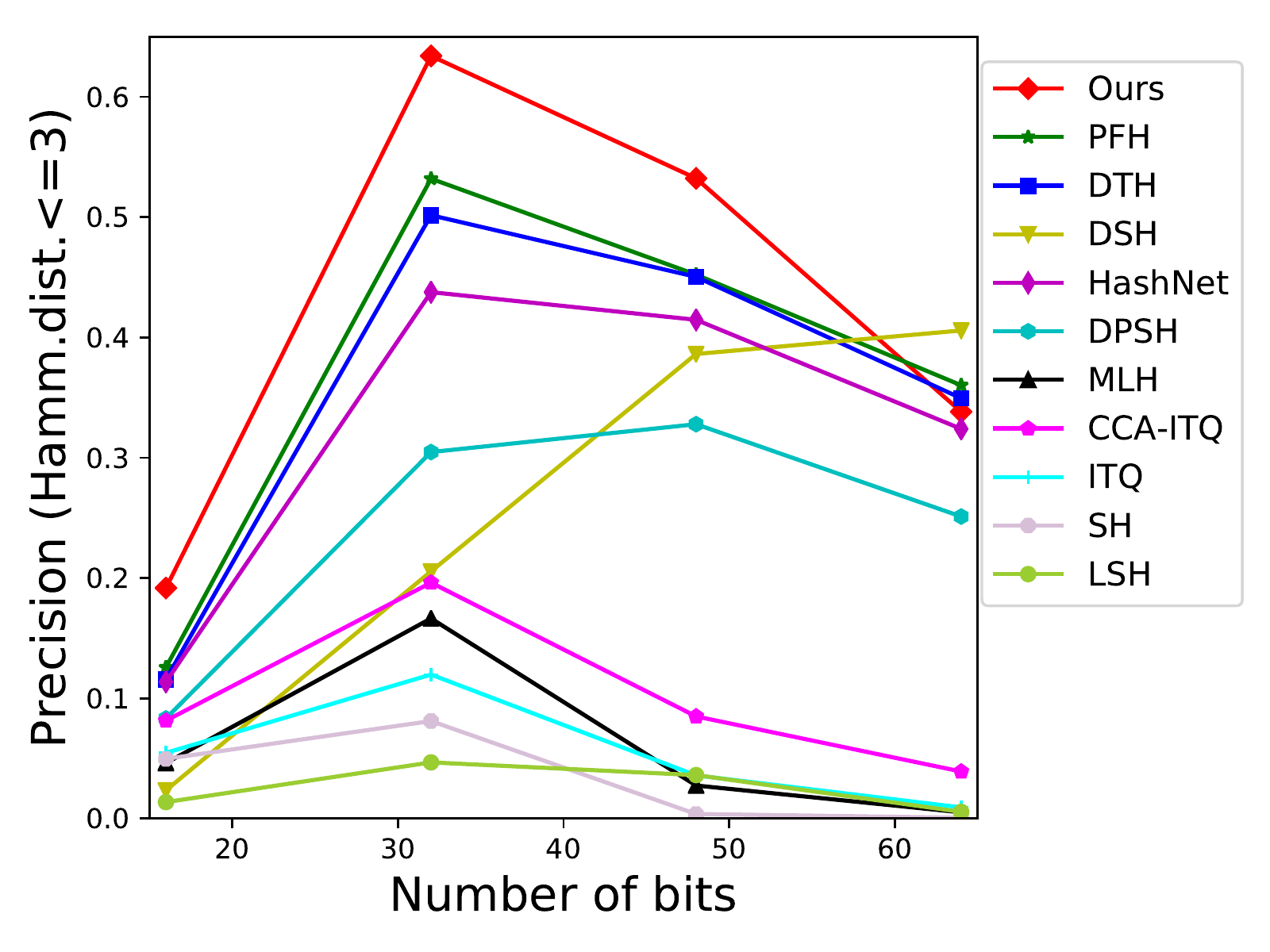}
}
\hspace{0.0in}
\subfigure[Precision-recall curve @ 16 bits]
{
\label{topp:subfig:b}
\includegraphics[width=2.15in]{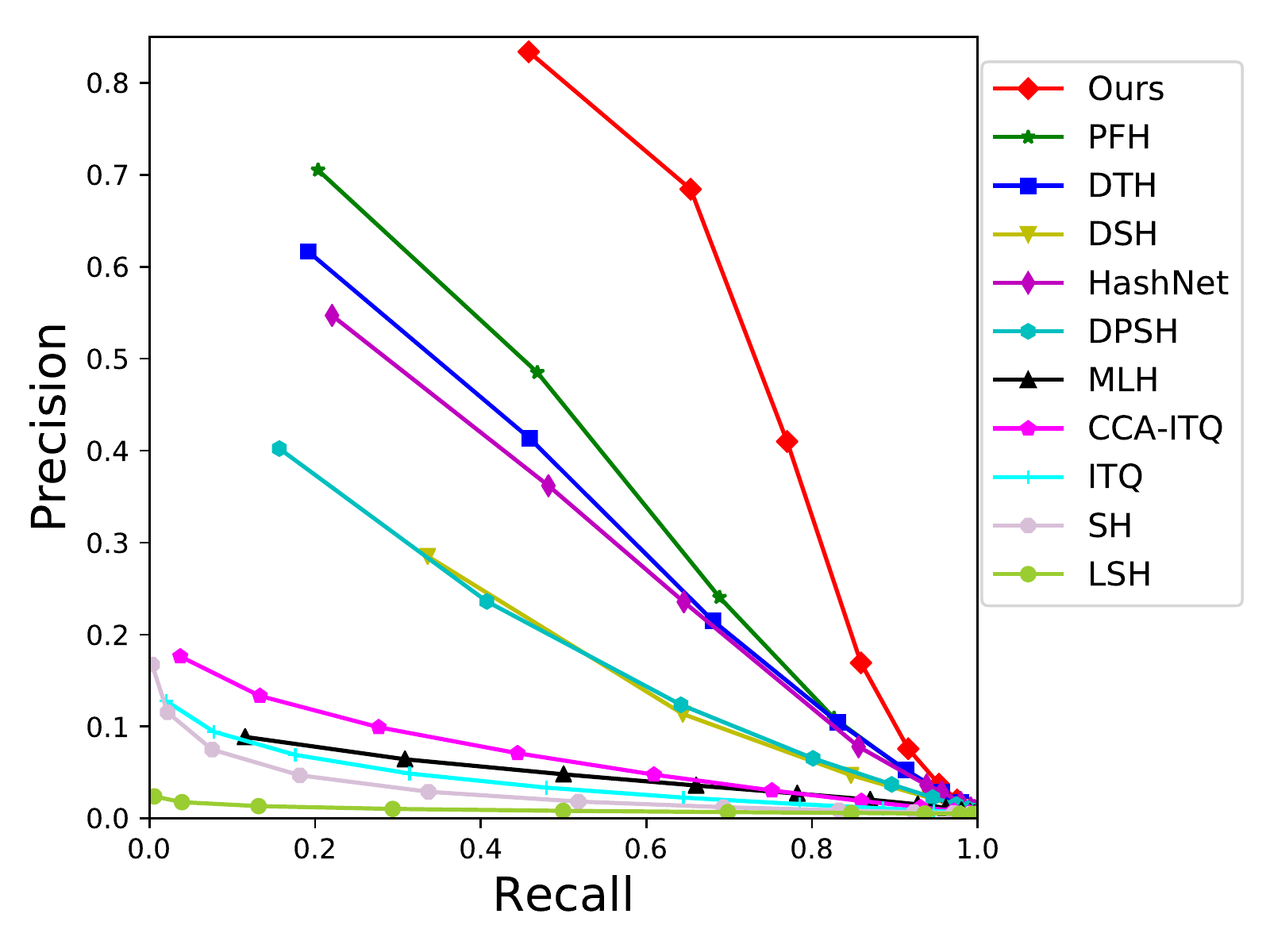}
}
\hspace{0.0in}
\subfigure[Precision curve w.r.t. top-N @ 16 bits]
{
\label{topp:subfig:b}
\includegraphics[width=2.15in]{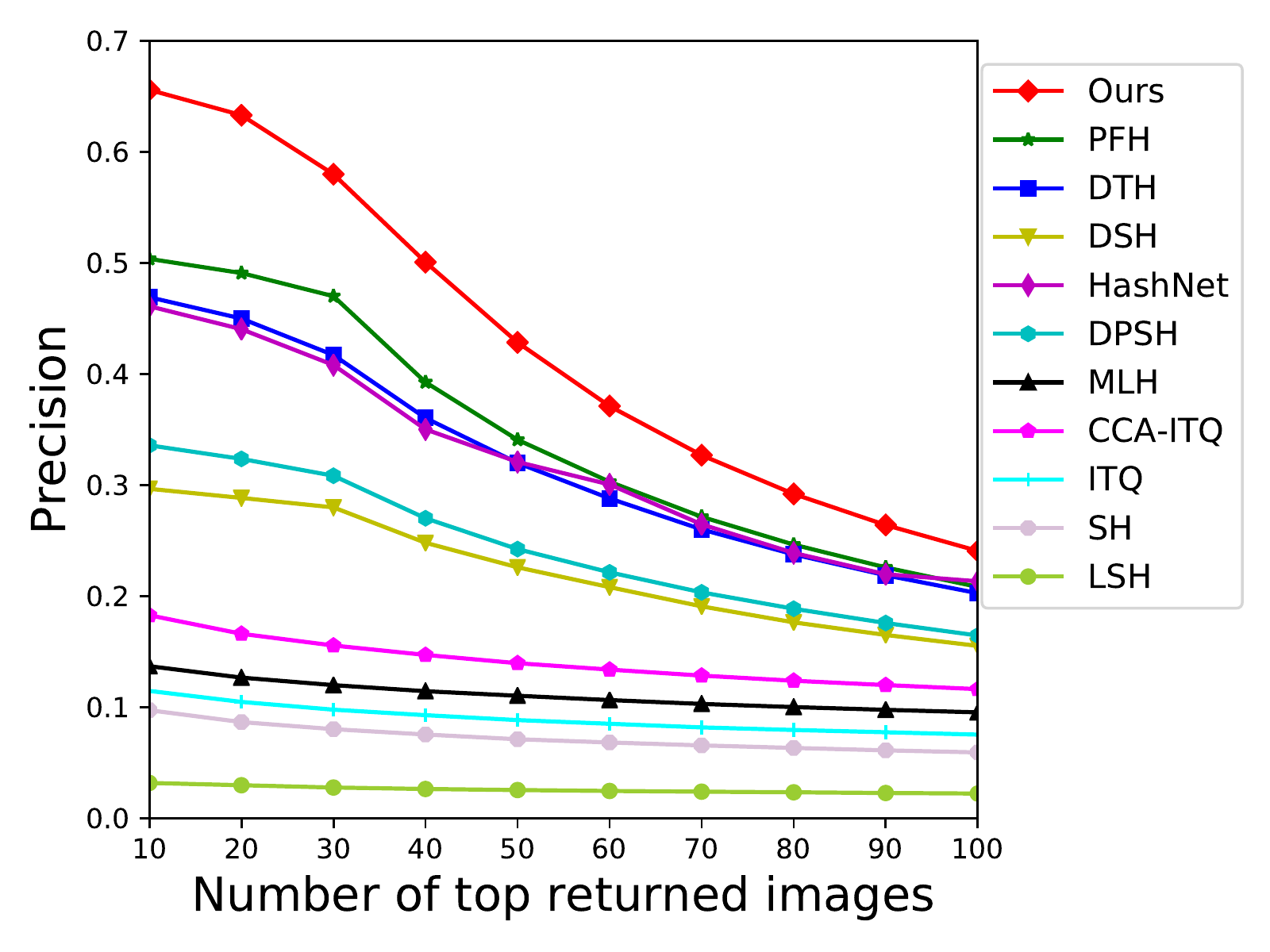}
}
\caption{The experimental results on the CUB-200-2011 dataset under three evaluation metrics.}
\label{topp1}
\end{figure*}

\begin{figure*}[ht!]
\centering
\subfigure[Precision within Hamming radius 3]
{
\label{topp:subfig:a}
\includegraphics[width=2.15in]{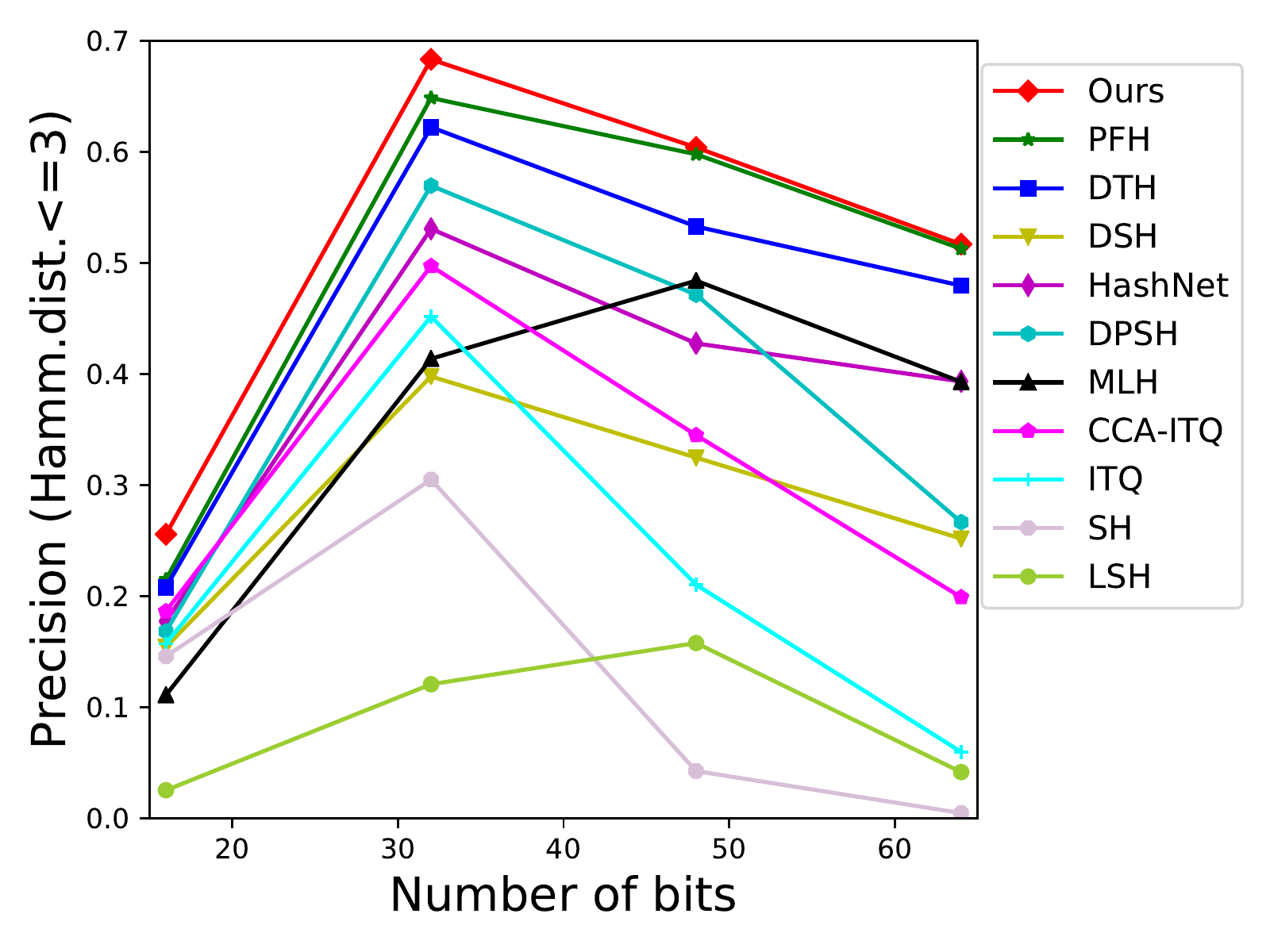}
}
\hspace{0.0in}
\subfigure[Precision-recall curve @ 16 bits]
{
\label{topp:subfig:b}
\includegraphics[width=2.15in]{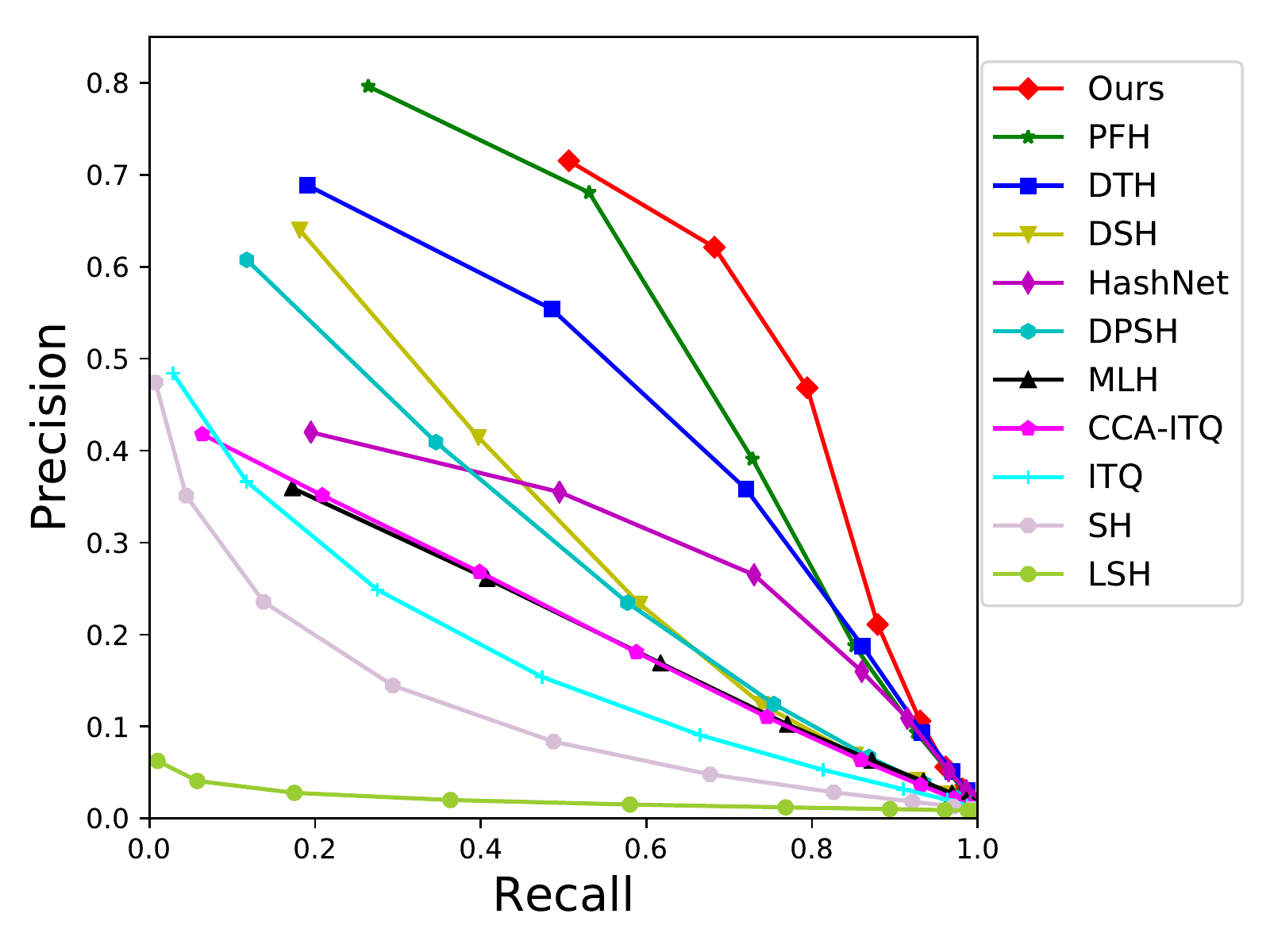}
}
\hspace{0.0in}
\subfigure[Precision curve w.r.t. top-N @ 16 bits]
{
\label{topp:subfig:b}
\includegraphics[width=2.15in]{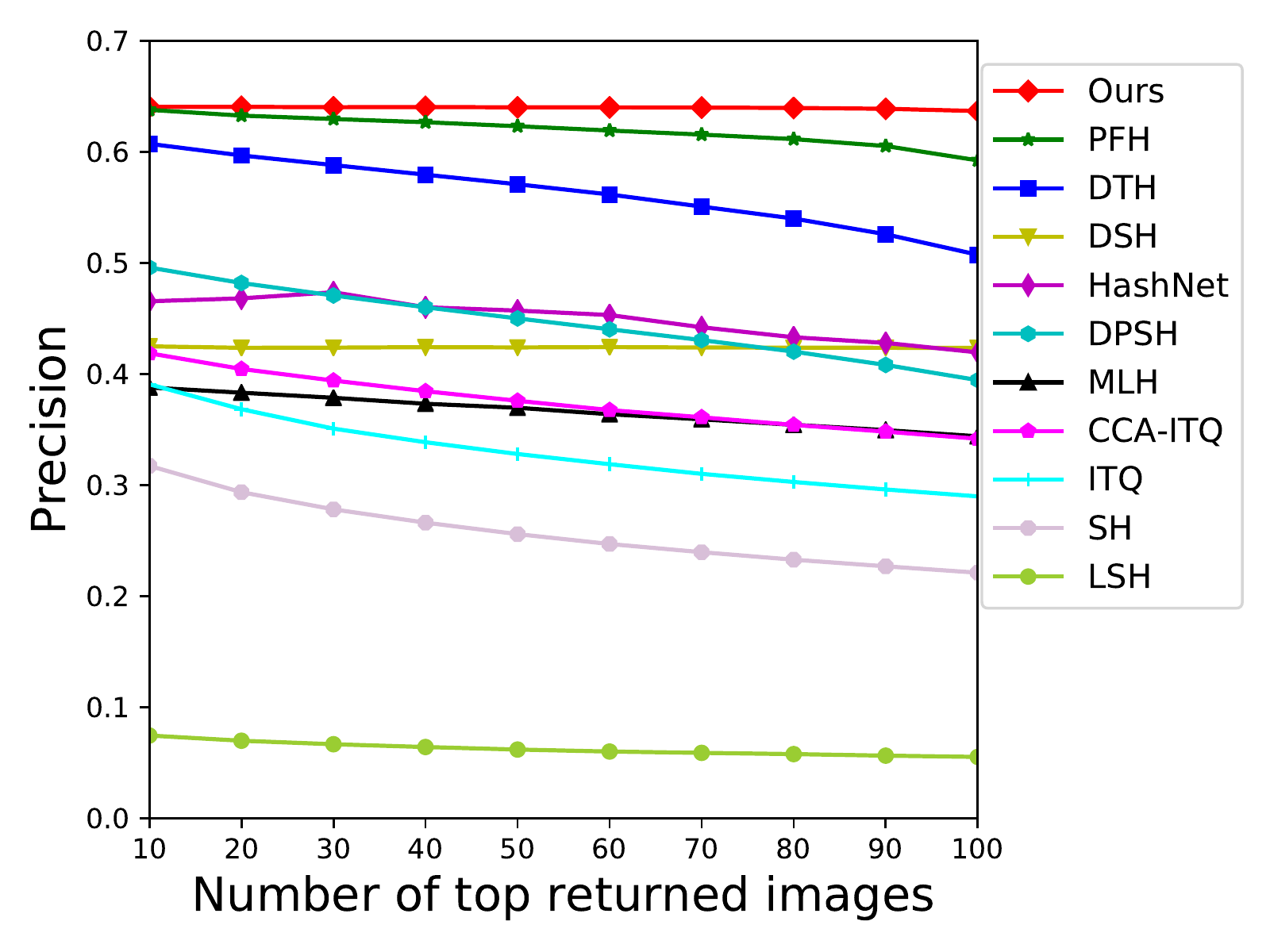}
}
\caption{The experimental results on the Standford Dogs dataset under three evaluation metrics.}
\label{topp2}
\end{figure*}

\subsection{Comparison with State-of-the-art Algorithms}
In the first set of experiments, we compare and evaluate the proposed method with several state-of-the-art baselines. 

In order to verify the effectiveness of our proposed method, 10 hashing methods are compared in the experiment. The baselines can be roughly divided into two categories: 1) the shallow methods and 2) the deep methods. The shallow methods use shallow architectures, e.g., linear model, to learn the hash functions. Iterative quantization with canonical correlation analysis (CCA-ITQ)~\cite{gong2013iterative}, minimal loss hashing (MLH)~\cite{norouzi2011minimal},  iterative quantization (ITQ)~\cite{gong2013iterative}, spectral hashing (SH)~\cite{weiss2009spectral} and locality-sensitive hashing (LSH)~\cite{gionis1999similarity} are belong to the shallow approaches. The deep methods use deep networks to learn more complex projections. Deep pairwise-supervised hashing (DPSH)~\cite{Li2016Feature}, deep supervised hashing~\cite{liu2016deep}, deep triplet hashing (DTH)~\cite{lai2015simultaneous}, HashNet~\cite{cao2017hashnet} and feature pyramid hashing (FPH)~\cite{FPH} are belong to the deep methods. The results of all baselines are directly cited from FPH. Note that both FPH and our method use the same deep architecture and are designed for fine-grained hashing.

\begin{table*}[htbp]
\centering
\caption{Comparison with DSaH and FPH of MAP on two fine-grained datasets.}
\label{MAP_Dash}
\begin{tabular}{|c|
p{1.2cm}<{\centering}p{1.2cm}<{\centering}p{1.2cm}<{\centering}p{1.2cm}<{\centering}|
p{1.2cm}<{\centering}p{1.2cm}<{\centering}p{1.2cm}<{\centering}p{1.2cm}<{\centering}|
}
\hline
\multirow{2}{*}{{Methods}}& \multicolumn{4}
{c|}{Oxford Flower-17}&\multicolumn{4}
{c|}{Stanford Dogs}\\
& 16bits & 32bits & 48bits & 64bits
& 16bits & 32bits & 48bits & 64bits
\\
\hline
{Ours}&  {\bf 0.9802}&  {\bf 0.9786}&  {\bf 0.9757}& {0.9753}& {\bf 0.6402}&  {\bf 0.7164}&  {\bf 0.7359}& {\bf 0.7401}\\ \hline
{FPH}&  {0.9542}&  {0.9653}&  {0.9691}& {\bf 0.9783}& {0.6224}&  {0.6688}&  {0.6924}& {0.6974}\\ \hline

DSaH &  0.9225&  0.9267&  0.9692&  0.9756&    0.3976&  0.5283&  0.5950&  0.6452\\  \hline

\end{tabular}
\end{table*}

Table~\ref{MAP} shows the comparison results of the MAP on the two fine-grained datasets. Figure~\ref{topp2} shows the curves of other three evaluation metrics. The results indicate that our method yields the highest accuracy and beats all baselines. Two observations can be made from the results. 

Firstly, compared with the coarse-grained approaches, our method performs significantly better than all previous baselines. Specifically, on CUB-200-2011, our method obtains a MAP of 0.6908 on 16 bits, compared with 0.4641 of DTH. On Stanford Dogs, our method obtains 0.7516, compared with 0.6362 of the second best coarse-grained method. The main reason is that all the coarse-grained methods are designed for the coarse-grained dataset. They aim to find the semantic differences but not the subtle differences. While, our method can localize the information regions and can encode the subtle differences into the binary codes.

Secondly, compared with the fine-grained approach, our method also performs significantly better than the FPH. For example, the MAP score of our method is 0.6922 when the bit length is 32, compared to 0.5832 of FPH on CUB-200-2011. On Stanford Dogs, the MAP of FPH is 0.6909, while our method can obtain 0.7402 on 32 bits. The main reason may be that the FPH only uses different layers to encode the binary codes while it can not accurately localize the informative regions. Different to that, our method can directly localize the informative regions. Also the proposed collaborative learning mechanism can learn the hash coding and localizer in reinforced way. 

\subsection{Comparison with Fine-grained Hashing Methods}
In the second set of experiments, we compare our method with other fine-grained hashing methods: DSaH~\cite{jin2018deep} and FPH.

Since the code of deep saliency hashing (DSaH) is not publicly available, we use the same experimental settings in DSaH for a fair comparison. DSaH, FPH and our method use the same settings. First,  the VGG-16~\cite{simonyan2014very} instead of ResNet-18 is used as the feature extractor. The VGG-16 is a deep architecture to learn the powerful image features. Second, we conduct experiments on an additional fine-grained dataset: Oxford Flower-17~\cite{Nilsback06}. The Oxford flower-17~\footnote{http://www.robots.ox.ac.uk/~vgg/data/flowers/17/} includes 17 flower species with 80 images for each class. The total number of images is 1360. For a fair comparison, we follow the setting in DSaH to split the data into training and test sets. All three methods use identical training and test sets.

The comparison results are shown in Table~\ref{MAP_Dash}. Again, our method yields the highest accuracy and beats all the baselines. For example, the MAP of our method is 0.9802 on 16 bits, compared with 0.9542 of FPH and 0.9225 of DSaH on Oxford Flower-17 dataset. On Stanford Dogs, our method obtains a MAP of 0.7401 on 64 bits, while only 0.6974 for FPH and 0.6452 for DSaH. The main reasons are: 1) DSaH trains the attention maps and hash coding separately, while our method train region localization and hash coding simultaneously; 2) FPH cannot find the discriminative regions while our method can.

\section{Conclusions}
In this work, we developed a fine-grained hashing method to capture the subtle differences among the fine-grained objects. We designed a novel architecture that can simultaneously train the region localization and hash code generation. These two tasks are corrected and can reinforce each other. The region localization module provides informative regions for the hash coding module and the hashing module provides supervised information to learn the localizer. Moreover, multi-scale regions were learned by utilizing multiple feature activations. Empirical evaluations on several fine-grained datasets showed that the proposed method achieves significantly better performance than the state-of-the-art fine-grained hashing baselines. In the future, we will study how to solve the large intra-class variation problem for fine-grained image retrieval.


%





\ifCLASSOPTIONcaptionsoff
  \newpage
\fi



%
%
%

\bibliographystyle{IEEEtran}
\bibliography{acmart}

\end{document}